\documentclass[letterpaper, 10 pt, conference]{ieeeconf}  

\usepackage{graphics} 
\usepackage{epsfig} 
\usepackage{mathptmx} 
\usepackage{times} 
\usepackage{amsmath} 
\usepackage{amssymb}  
\usepackage{hyperref}
\usepackage{diagbox}  
\usepackage{multirow} 
\usepackage{siunitx}
\usepackage{makecell}

\usepackage{diagbox}  
\usepackage{multirow} 

\title{\LARGE \bf
Adaptive MPC-based quadrupedal robot control under periodic disturbances
}

\author{Elizaveta Pestova, Ilya Osokin, Danil Belov, Pavel Osinenko
}

\begin{document}

\maketitle
\thispagestyle{empty}
\pagestyle{empty}

%
\begin{abstract}


Recent advancements in adaptive control for reference trajectory tracking enable quadrupedal robots to perform locomotion tasks under challenging conditions. There are methods enabling the estimation of the external disturbances in terms of forces and torques. However, a specific case of disturbances that are periodic was not explicitly tackled in application to quadrupeds. This work is devoted to the estimation of the periodic disturbances with a lightweight regressor using simplified robot dynamics and extracting the disturbance properties in terms of the magnitude and frequency. Experimental evidence suggests performance improvement over the baseline static disturbance compensation.

All source files, including simulation setups, code, and calculation scripts, are available on GitHub at \url{https://github.com/LizaP9/Periodic_Adaptive_MPC}.

\end{abstract}


\section{INTRODUCTION}

Quadrupedal robots often operate in dynamic and unpredictable environments, where they are subjected to external disturbances and unmodeled forces, such as payloads and environmental interactions.
Traditional Model Predictive Control (MPC) approaches struggle with these uncertainties, as they rely on an accurate system model \cite{convex_MPC}.
To address these challenges, various adaptive control techniques and force estimation strategies have been proposed in the literature.

A significant advancement in adaptive MPC is the incorporation of Control Lyapunov Functions (CLF) to 
enforce stability constraints within the optimization problem \cite{AdaptiveCLF_MPC}, \cite{KrsticCLF}. 
This approach, referred to as Adaptive CLF-MPC, introduces an additional term in the cost function that 
penalizes deviations from a Lyapunov-stable trajectory, ensuring stability even in the presence of unknown 
disturbances \cite{SlotineAppliedNonlinearControl}. This framework has been successfully validated on quadrupedal robots, demonstrating improved 
robustness when handling payload-induced forces \cite{SlotineAdaptiveControl}, \cite{AdaptiveMetaLearning}.

To ensure safe operation under uncertain conditions, adaptive safety with control barrier functions has also been proposed \cite{TaylorAdaptiveSafetyControl}.
Furthermore, Bayesian learning-based adaptive control techniques have shown promise in safety-critical applications \cite{FanBayesianAdaptiveControl}.

Another approach involves integrating a force adaptation mechanism within the MPC framework \cite{HighlyDynamicMPC}. 
This method modifies the system dynamics to account for estimated external forces, thereby improving the 
robot's ability to compensate for disturbances in real-time. The adaptive formulation enables continuous 
refinement of the force predictions based on sensory feedback, leading to more stable locomotion.
For example, in \cite{DingMPC}, a real-time MPC approach was developed to enable versatile and 
dynamic motions in quadrupedal robots.
The proposed controller demonstrated the ability to handle rapid changes in gait and adapt to 
challenging terrains, highlighting the effectiveness of MPC in dynamic and unpredictable environments.

For systems with dynamic environments, nonlinear model predictive control methods combined with control Lyapunov functions can further enhance robustness \cite{GrandiaNonlinearMPC}.

\begin{figure}[t]
    \centering
    \includegraphics[width=0.9\linewidth]{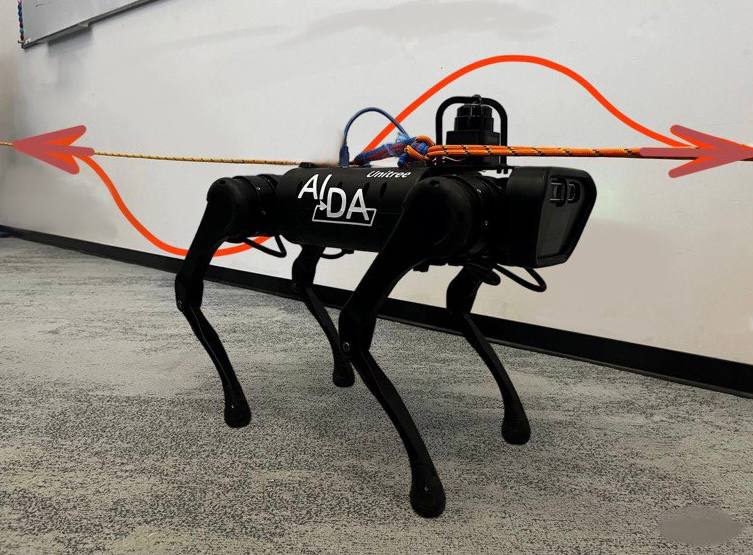}
    \caption{A quadrupedal robot under periodic disturbances}
    \label{fig:dog_image}
\end{figure}

Furthermore, research has explored combining Whole-Body Control (WBC) with MPC \cite{NonlinearMPC}, \cite{GuidedCPO}. 
In this hybrid control strategy, MPC computes optimal reaction forces over a prediction horizon, while WBC refines these 
forces by considering full-body dynamics and joint-level constraints. This integration has been shown to enhance 
the agility and stability of quadrupedal robots, particularly in high-speed locomotion scenarios.
Contact-implicit optimization methods \cite{ContactImplicit_MPC} have been explored to improve 
computational efficiency by eliminating the need for explicit force modeling in dynamic environments.

Nonlinear Model Predictive Control (NMPC) has been explored as an alternative to standard MPC for handling 
parameter uncertainty in robotic systems \cite{NonlinearMPC}. NMPC leverages 
online optimization techniques, such as Sequential Quadratic Programming (SQP), to iteratively adjust 
force compensations in real time. While NMPC provides superior tracking accuracy under uncertainties, 
its high computational complexity limits its feasibility for real-time applications on quadrupedal robots \cite{MetaStrategyOptimization}.

To address the computational challenges of NMPC, researchers have investigated hybrid approaches that 
integrate learning-based parameter estimation \cite{HybridRL_MPC}. Gaussian 
Process Regression (GPR) models, for example, have been employed to predict external force patterns 
based on past observations. Although these methods improve robustness, they require large datasets for 
training and may not generalize well to novel disturbances. For instance, an online learning framework 
has been proposed to model unknown dynamics
in legged locomotion, enabling more accurate force estimation and adaptive control in real-time 
scenarios \cite{SunOnlineLearning}.

Recent advances suggest that hybrid control architectures, combining MPC with Reinforcement Learning (RL) and Imitation Learning (IL), 
offer promising solutions for adaptive locomotion \cite{Hybrid_Imitation_MPC}. In such 
frameworks, RL-based policies learn to optimize force compensations, while MPC ensures constraint satisfaction 
and stability \cite{JoshiDeepModelReferenceAdaptiveControl}. This approach enables quadrupedal robots to adapt to varying terrain conditions and dynamic 
external forces.

For instance, a hierarchical control scheme has been proposed in which a high-level RL controller selects 
optimal contact configurations, while a low-level MPC controller computes reaction forces \cite{HybridRL_Hierarchy}. 
This formulation allows for real-time adaptation to external disturbances without requiring extensive model re-tuning.

$L_1$ adaptive control strategies have been employed successfully in bipedal robotics, demonstrating how control Lyapunov 
functions can be combined with quadratic programming to improve stability and performance \cite{L1AdaptiveControlBipedal}.

Accurate force estimation is critical for adaptive control and disturbance rejection in quadrupedal robots \cite{AdaptiveForceMPC}. 
Traditional force estimation methods rely on inverse dynamics calculations, which require precise knowledge 
of the system model. However, these methods struggle when unmodeled forces, such as payload variations or 
terrain interactions, are present.

A widely used technique for force estimation is Dynamic Regressor Extension and Mixing 
(DREM) \cite{DREM}. DREM enhances parameter estimation accuracy by 
transforming the system equations into a form that allows independent estimation of each parameter. 
Unlike classical Least-Squares (LS) estimators, which require persistent excitation conditions for 
convergence, DREM achieves faster and more reliable parameter estimation under a broader range of conditions.

An extension of DREM, known as Composite Adaptive Disturbance Rejection, further improves force estimation 
robustness by incorporating instrumental variable regression techniques \cite{NewResultsDREM}. 
This method reduces sensitivity to measurement noise and provides more accurate estimates of external wrenches 
acting on the robot's base.

Additionally, implicit regularization techniques combined with momentum algorithms have been explored
to improve parameter estimation in nonlinearly parameterized adaptive control systems, leading to enhanced
stability and prediction accuracy \cite{BoffiSlotine2021}.

The reviewed studies highlight the limitations of standard MPC in dealing with external forces and demonstrate 
the effectiveness of adaptive and force-estimation-based enhancements. Approaches such as Adaptive CLF-MPC, 
Composite Adaptive Control, and DREM-based force estimation have shown promising results in improving 
quadrupedal locomotion robustness and stability. A novel composite adaptation scheme using the Dynamic Regressor 
Extension and Mixing (DREM) 
method was proposed to ensure asymptotic parameter estimation and reference tracking in the presence of external 
torques, while relaxing the persistency of excitation requirement \cite{GlushchenkoCompositeDREM}. Future research 
should focus on real-time implementation 
on hardware.

Recent advancements in adaptive control of quadrupedal robots allow for accurate static disturbance estimation.
On the other hand, there are numerous techniques in the estimation of different kinds of disturbances, including the general case and periodic disturbances as well.
However, to our knowledge there were no prior work focusing on applying MPC-based compensation to a quadrupedal robot with explicit evaluation of periodic disturbances.

In this paper, the case of a quadrupedal robot subject to external disturbances was addressed, as illustrated in Fig.~\ref{fig:dog_image}.
The contributions of this work are as follows.

First, a lightweight regression-based disturbance estimation was implemented based on the simplified dynamics of a quadrupedal robot.
Second, a disturbance separation into stationary and non-stationary parts was realized.

Third, adaptation was added to the control system of the robot on top of the existing MPC-based control algorithm.

Finally, a number of experiments in the simulation were conducted with varying periodical disturbances on a Unitree A1 quadrupedal platform.
Quality of the reference tracking for the proposed algorithm versus the baseline was evaluated.
The results suggest that adding periodic disturbances compensation improves the reference tracking quality in comparison with the stationary disturbance compensation.

\section{System Dynamics and Control}

\subsection{Robot Dynamics Model}

\subsubsection{Floating-Base Dynamics}

Quadrupedal robots are underactuated systems that interact with their environment through contact points,
making their dynamics fundamentally different from wheeled or fixed-base robots. Unlike robotic manipulators,
that are fixed to a base, quadrupedal robots have a floating base, meaning their center of mass (CoM) is
not directly actuated and can move freely in 3D space. This requires accounting for the full rigid body dynamics, 
including both linear and angular motion, while ensuring the feasibility of the contact forces.

To accurately model these dynamics, floating-base dynamics is used, and the robot is treated as a rigid body
influenced by the contact forces at its feet. This approach allows for accurate modeling of the ground reaction forces (GRFs)
and external disturbances, which are crucial for adaptive MPC control strategies.

\subsubsection{Equations of Motion}

The dynamics of a quadrupedal robot are governed by Newton-Euler equations, that describe the evolution of
translational and rotational motion:

\begin{equation}
   m \ddot{\mathbf{p}} = \sum_{i=1}^{n_c} \mathbf{f}_i - m \mathbf{g},
   \label{eq:linear_dynamics}
\end{equation}

\begin{equation}
   \frac{d}{dt} (\mathbf{I} \boldsymbol{\omega}) = \sum_{i=1}^{n_c} \mathbf{r}_i \times \mathbf{f}_i,
   \label{eq:angular_dynamics}
\end{equation}

where:
\( m \) is the total mass of the robot,
\( \mathbf{p} \in \mathbb{R}^3 \) is the position of the center of mass (CoM) in the world frame,
\( \mathbf{g} \) is the gravitational acceleration vector,
\( n_c \) is the number of active contact points,
\( \mathbf{f}_i \in \mathbb{R}^3 \) is the ground reaction force (GRF) exerted at each contact point \( i \),
\( \mathbf{I} \in \mathbb{R}^{3 \times 3} \) is the inertia matrix of the robot (expressed in the body frame),
\( \boldsymbol{\omega} \in \mathbb{R}^3 \) is the angular velocity of the floating base,
\( \mathbf{r}_i \) is the translation vector from the CoM to the \( i \)-th contact point.

The first equation describes the linear acceleration of the robot's center of mass, which results from the sum of 
all external forces, including contact forces and gravity. The second equation represents the angular acceleration,
which is influenced by external torques generated from the contact forces.

\subsubsection{State Representation with External Forces}

In real-world scenarios, the system is affected by unknown external forces, which can lead to
significant discrepancies between the expected and actual robot behavior, necessitating a 
force estimation mechanism to improve the control robustness.

To address this, a vector of unknown external parameters is introduced, denoted as:

\begin{equation}
   \boldsymbol{\xi} = \begin{bmatrix} \mathbf{f}_{unk} \\ \mathbf{t}_{unk} \end{bmatrix} \in \mathbb{R}^6,
\end{equation}

where:
\( \mathbf{f}_{unk} \in \mathbb{R}^3 \) is the compensation vector for the unknown external force acting on the robot’s CoM,
\( \mathbf{t}_{unk} \in \mathbb{R}^3 \) is the compensation vector for the unknown external torque affecting the robot’s rotational dynamics.



The state vector of the robot is as follows:

\begin{equation}
   \mathbf{x} = \begin{bmatrix}
      \boldsymbol{\theta} &
      \mathbf{p} & 
      \boldsymbol{\omega} &
      \mathbf{v} 
   \end{bmatrix}^T,
   \label{eq:state_vector}
\end{equation}

where \( \boldsymbol{\theta} \in \mathbb{R}^3 \) represents the orientation of the floating base (Euler angles or a rotation matrix),
\( \mathbf{v} = \dot{\mathbf{p}} \in \mathbb{R}^3 \) is the linear velocity of the CoM, and \( \boldsymbol{\omega} \) is the angular velocity.

Thus, the floating-base dynamics of the quadrupedal robot, considering the unknown external forces and torques \cite{AdaptiveCLF_MPC}, 
can be written as:

\begin{equation}
   \begin{cases}
      \dot{\boldsymbol{\theta}} = T(\boldsymbol{\theta}) \boldsymbol{\omega}, \\[5pt]
      \dot{\mathbf{p}} = \mathbf{v}, \\[5pt]
      \dot{\boldsymbol{\omega}} = \mathbf{I}^{-1} \left( -\boldsymbol{\omega} \times \mathbf{I} \boldsymbol{\omega} + \sum_{i=1}^{n_c} \mathbf{r}_i \times \mathbf{f}_{i} - \mathbf{t}_{unk} \right), \\[5pt]
      \dot{\mathbf{v}} = \mathbf{g} + \frac{1}{m} \left( \sum_{i=1}^{n_c} \mathbf{R}_{wb} \boldsymbol{f}_{i} - \mathbf{f}_{unk} \right)
   \end{cases}
   \label{eq:state_evolution_with_uncertainty}
\end{equation}

where:
\( T(\boldsymbol{\theta}) \) is the transform matrix representing angular velocity in the Euler angles form,
\( \mathbf{R}_{wb} \) is the rotation matrix converting forces from the body frame to the world frame.

\subsection{Model Predictive Control Formulation}

A baseline controller that the proposed method is built upon is formulated as follows.

The objective of the Model Predictive Controller (MPC) is to compute the ground reaction forces that enable 
the quadrupedal robot to follow a given trajectory while maintaining balance. Instead of directly controlling 
joint torques, the optimization is performed at the force level, simplifying the problem by focusing on the body
dynamics rather than low-level joint configurations.

At each iteration, MPC predicts the evolution of the system state over a finite horizon and finds such 
control inputs that minimize a cost function while satisfying dynamic and physical constraints. The 
optimization is performed at each time step, only the first control input from the sequence is applied, and 
the process is repeated.

In order to speed up the computations, a convex optimization problem is formulated. The contact sequence (i.e., stance and swing phases) are generated prior to MPC on the level of the gait scheduler and step planner, ensuring that the MPC problem remains convex and converges to a unique global minimum.

The discrete-time MPC problem is formulated as an optimal control problem:

\begin{equation}
    \min_{\mathbf{u}} \sum_{i=0}^{k-1} 
    \left( \| \mathbf{x}_{i+1} - \mathbf{x}_{i+1}^{\text{ref}} \|_{\mathbf{P}}^2 
    + \| \mathbf{u}_i \|_{\mathbf{R}}^2 \right),
    \label{eq:mpc_cost_function}
\end{equation}
subject to system dynamics:
\begin{equation}
    \mathbf{x}_{i+1} = \mathbf{A}_i \mathbf{x}_i + \mathbf{B}_i \mathbf{u}_i, 
    \quad i = 0, \dots, k-1,
    \label{eq:mpc_dynamics}
\end{equation}
input constraints:
\begin{equation}
    \mathbf{c}_{\min} \leq \mathbf{C}_i \mathbf{u}_i \leq \mathbf{c}_{\max}, 
    \quad i = 0, \dots, k-1,
    \label{eq:mpc_constraints}
\end{equation}
and gait-specific force constraints:
\begin{equation}
    \mathbf{D}_i \mathbf{u}_i = 0, 
    \quad i = 0, \dots, k-1.
    \label{eq:mpc_gait_constraints}
\end{equation}

where:
\( \mathbf{x}_i \), \( \mathbf{u}_i \) is the system state and control inputs (GRF) at time step \( i \),
\( \mathbf{x}_{i+1}^{\text{ref}} \) is the desired reference state at time step \( i+1 \),
\( \mathbf{P} \) and \( \mathbf{R} \) are positive semi-definite weight matrices penalizing deviations from the reference trajectory and control effort,
\( \mathbf{A}_i \) and \( \mathbf{B}_i \) define the discrete-time system dynamics,
\( \mathbf{C}_i \) and \( \mathbf{c} \) define inequality constraints on the control inputs,
\( \mathbf{D}_i \) ensures that the ground reaction forces from the feet that are not in contact with the ground are set to zero.

\subsection{System Linearization and Discretization}

Even though the floating-base dynamics is nonlinear due to the cross-product terms in the momentum equation and 
the rotational kinematics, the MPC problem remains convex through the linearization of the system.

\begin{figure}[h]
    \centering
    \includegraphics[width=1.0\linewidth]{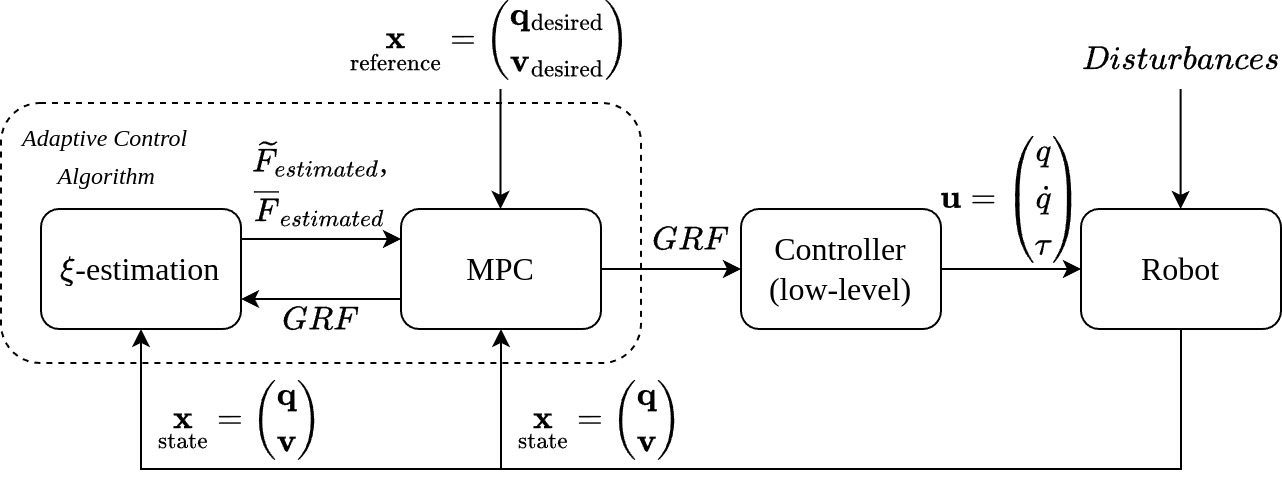}
    \caption{Block diagram of the adaptive MPC-based control system}
    \label{fig:scheme}
\end{figure}

Figure~\ref{fig:scheme} illustrates the high-level control architecture. 
The baseline MPC computes ground reaction forces under a simplified model assumptions. 
An external disturbance estimator continuously refines the dynamical model by estimating the unknown forces.

To maintain convexity, the following approximations and simplifications are used.
\begin{itemize}
    \item Small Angle Assumption: The roll and pitch angles of the body are assumed to be small, simplifying the coordinate transforms.
    \item Time-Varying Linearization: The torque is computed based on the commanded limb configuration.
    \item Neglecting Off-Diagonal Inertia Terms: The inertia tensor is approximated as a diagonal matrix.
    \item The quadratic term \(\boldsymbol{\omega} \times \mathbf{I} \boldsymbol{\omega}\) is neglected.
\end{itemize}

Under these assumptions, the discrete-time system dynamics used in the MPC formulation are:

\begin{equation}
   \mathbf{x}_{k+1} = \mathbf{A}_d \mathbf{x}_k + \mathbf{B}_d \mathbf{u}_k + \mathbf{G}_d + \mathbf{Q}_d \boldsymbol{\xi}_k,
   \label{eq:mpc_discrete_dynamics}
\end{equation}

where the system matrices are:

\begin{equation}
    \mathbf{A}_d =
    \begin{bmatrix}
        \mathbf{1}_{3\times3} & \mathbf{0}_{3\times3} & T(\boldsymbol{\theta}) \Delta t & \mathbf{0}_{3\times3} \\
        \mathbf{0}_{3\times3} & \mathbf{1}_{3\times3} & \mathbf{0}_{3\times3} & \mathbf{1}_{3\times3} \Delta t \\
        \mathbf{0}_{3\times3} & \mathbf{0}_{3\times3} & \mathbf{1}_{3\times3} & \mathbf{0}_{3\times3} \\
        \mathbf{0}_{3\times3} & \mathbf{0}_{3\times3} & \mathbf{0}_{3\times3} & \mathbf{1}_{3\times3}
    \end{bmatrix},
\end{equation}

\begin{equation}
    \mathbf{B}_d =
    \begin{bmatrix}
        \mathbf{0}_{3\times3} & \dots & \mathbf{0}_{3\times3} \\
        \mathbf{0}_{3\times3} & \dots & \mathbf{0}_{3\times3} \\
        \mathbf{I}^{-1} [\mathbf{r}_1]_\times \Delta t & \dots & \mathbf{I}^{-1} [\mathbf{r}_{n_c}]_\times \Delta t \\
        \mathbf{1}_{3\times3} \Delta t / m & \dots & \mathbf{1}_{3\times3} \Delta t / m
    \end{bmatrix},
\end{equation}

\begin{equation}
    \mathbf{Q}_d =
    \begin{bmatrix}
        \mathbf{0}_{6\times6} \\
        \mathbf{1}_{6\times6} \\
    \end{bmatrix},
 \end{equation}

\begin{equation}
    \mathbf{G}_d =
    \begin{bmatrix}
       0 \ 0 \ 0 \ 0 \ 0 \ 0 \ 0 \ 0 \ 0 \ 0 \ 0 \    \mathbf{g}
    \end{bmatrix}^T.
 \end{equation}
 


\section{ External Force Estimation}

The estimate of the unknown external forces acting on the quadrupedal robot is used to make the system model for MPC more accurate. Initially, MPC operates with a nominal model that does not explicitly account for unknown disturbances. But as the system evolves, the discrepancies between the predicted and actual states are used to evaluate them.

The estimation problem is stated as follows. Given the real state trajectory of the robot, that was observed after applying the control inputs computed by MPC, and the expected trajectory calculated with the nominal system model, estimate the unknown forces \( \boldsymbol{\xi} \) that cause the discrepancies between these trajectories.






In order to find this vector, an optimization problem is formulated.
An optimal estimate \( \boldsymbol{\xi}^* \) is said to minimize the quadratic cost function:


\begin{equation}
    \min_{\boldsymbol{\xi}} \left\| \mathbf{x}_{k+1}^{\text{real}} - \mathbf{A}_d \mathbf{x}_k^{\text{real}} - \mathbf{B}_d \mathbf{u}_k - \mathbf{Q}_d \boldsymbol{\xi} - \mathbf{G}_d \right\|^2_{\mathbf{S}}
    \label{eq:minimization_problem}
\end{equation}
 
where \( \mathbf{S} \in \mathbb{R}^{6 \times 6} \) is a weighting matrix that penalizes the estimation errors.

The closed-form solution to this problem reads:


\begin{equation}
    \boldsymbol{\xi}^* = \left( \mathbf{Q}_d^T \mathbf{S} \mathbf{Q}_d \right)^{-1} \mathbf{Q}_d^T \mathbf{S} (\mathbf{x}_{k+1}^{\text{real}} - \mathbf{A}_d \mathbf{x}_k^{\text{real}} - \mathbf{B}_d \mathbf{u}_k - \mathbf{G}_d).
    \label{eq:optimal_estimation}
\end{equation}



After the forces are estimated, they are used in the modified dynamics model of the system, thereby allowing for the control that takes into account the external disturbances.

   


\section{Experimental Setup}

\subsection{Simulation Environment}

Experiments were conducted in the Raisim simulator using a Unitree A1 quadrupedal robot. The simulator is fully integrated with ROS for real-time control and sensor feedback.

\begin{figure}[h]
    \centering
    \includegraphics[width=0.49\linewidth, height=0.25\linewidth]{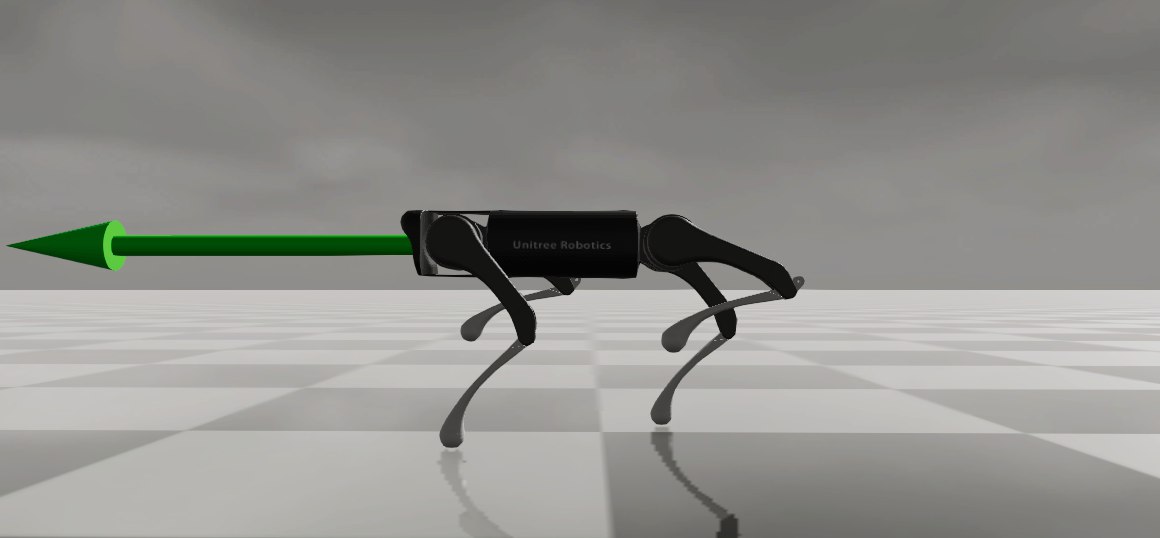}
    \hfill
    \includegraphics[width=0.49\linewidth, height=0.25\linewidth]{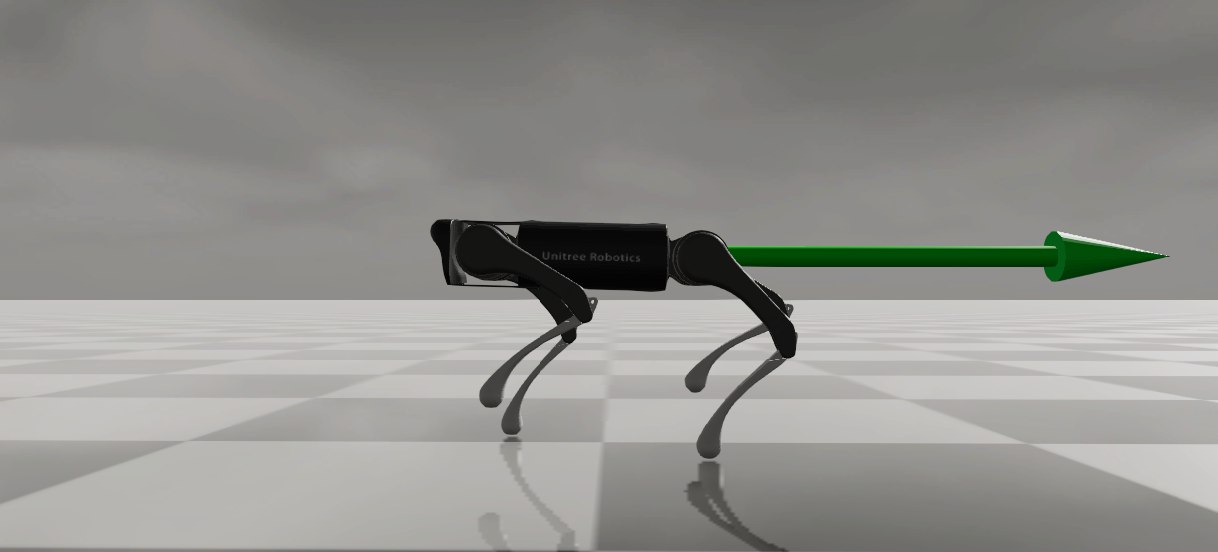}
    \caption{Simulation setup in Raisim}
    \label{fig:simulation_setup}
\end{figure}

Figure~\ref{fig:simulation_setup} figure shows two screenshots from the simulation, where A1 walks under the influence of a periodically changing disturbance.

\subsection{Testing Scenarios}

The experiments were conducted with periodical external disturbances of varying amplitudes and static components,
assessing three distinct robot behaviors: static balance (zero velocity), 
low-speed trotting, and high-speed trotting.

The simulated external forces are as follows:
\begin{equation}
    \text{Disturbance} = d_{\text{static}} + A \cdot \sin(\omega t)
\end{equation}

\subsection{Performance Metrics}

The reference tracking quality was evaluated using the mean squared error (MSE) between the desired and measured trajectories.

\begin{equation}
    \sum_{k=1}^{N} \left\| \mathbf{x}_{k}^{\text{reference}} - \mathbf{x}_k^{\text{real}} \right\|^2_{\mathbf{S}}.
    \label{eq:reference_tracking_metric}
 \end{equation}

 \section{Results and Discussion}

 \subsection{Quantitative Results}
\begin{figure}[h]
     \centering
     \includegraphics[width=0.95\linewidth]{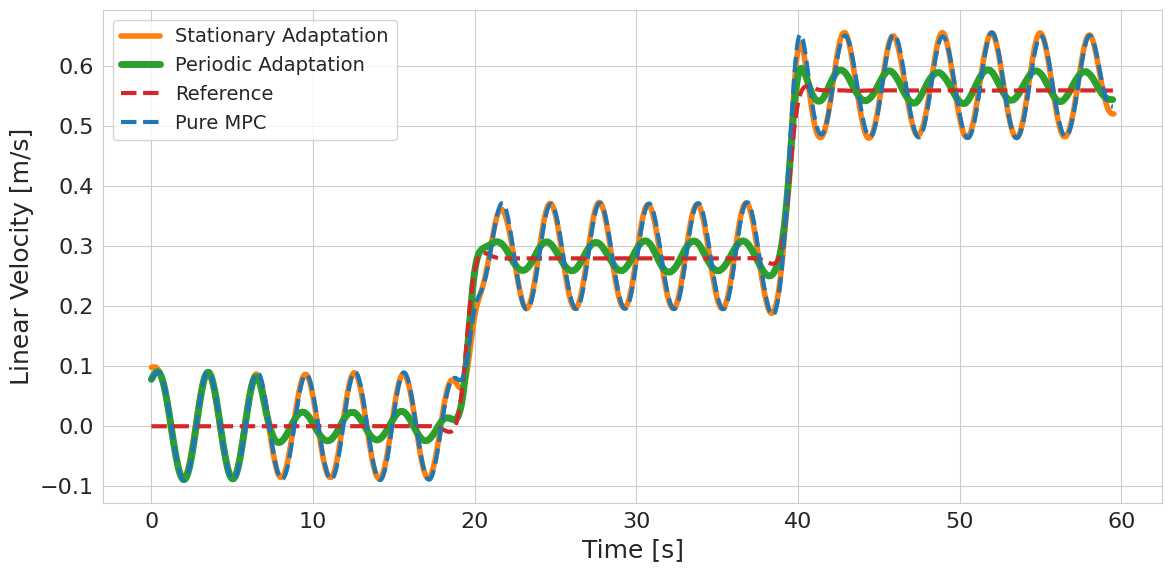}
     \caption{Comparison of commanded (dashed red) and measured linear velocities for different control strategies under a periodic disturbance.}
     \label{fig:velocity_comparison}
 \end{figure}

 The experimental results are summarized in Table~\ref{table:metrics} and Fig.~\ref{fig:velocity_comparison}, where the commanded and measured velocities are compared under different disturbance conditions. Each plot is divided into three segments corresponding to zero, low, and high linear velocities, highlighting the robot's performance across various operating regimes. Three control strategies were evaluated:
 
 \begin{itemize}
     \item \textbf{No Compensation (Baseline MPC)} uses the nominal model without disturbance estimation.
     \item \textbf{Static Disturbance Compensation} uses a static estimate computed over a prolonged time window.
     \item \textbf{Periodic Compensation (Ours)} separately estimates and compensates for the static and periodic components of the disturbance.
 \end{itemize}
 
 The tracking quality was measured using the mean squared error (MSE) between the reference and measured trajectories, as defined in~\eqref{eq:reference_tracking_metric}. The numerical results for various disturbance scenarios are presented in Table~\ref{table:metrics}.

\begin{table}[ht]
    \centering
    \caption{Tracking performance ($1000 \times$ MSE for $X$ velocity) 
             under various disturbance scenarios.}
    \label{table:metrics}
    \resizebox{\columnwidth}{!}{
    \begin{tabular}{|c|c|c|c|c|c|}
        \hline
        \multirow{2}{*}{%
          \begin{minipage}[c][\height][c]{1.2cm}
            \centering
            \noindent
            Frequency [Hz]
          \end{minipage}%
        } 
        & \multicolumn{2}{|c|}{Disturbances [N]} 
        & \multicolumn{2}{c|}{Compensation} 
        & \multirow{2}{*}{\makecell{Baseline \\ MPC}} 
        \\
        \cline{2-5} 
         & Stationary & \makecell{Periodic \\ (Amplitude)} & Static & Periodic &  \\
        \hline
        0.33 & 0   & 15 & 5.663 & \textbf{2.690} & 6.471 \\
        \hline
        0.33 & 0   & 10 & 2.755 & \textbf{1.895} & 3.676 \\
        \hline
        0.33 & -10 & 0  & 1.261 & 1.290 & \textbf{1.222} \\
        \hline
        0.33 & -7  & 10 & 3.175 & \textbf{1.819} & 3.567 \\
        \hline
        0.33 & -10 & 15 & 6.035 & \textbf{2.728} & 6.694 \\
        \hline
        0.5  & -10 & 15 & 15.351 & \textbf{4.264} & 15.468 \\
        \hline
    \end{tabular}}
\end{table}

 Table~\ref{table:metrics} compares the MSE for each disturbance scenario across the three methods. Lower values indicate more accurate velocity tracking. As it could be seen, the proposed periodic compensation method (fourth column) yields the lowest tracking error when the disturbances have a periodic component.
 
 Figure~\ref{fig:velocity_comparison} shows a velocity profile over time. The robot experiences significant oscillations with purely stationary compensation (orange) and the baseline MPC without compensation (blue). In contrast, the proposed periodic compensation approach (green) closely follows the reference trajectory, mitigating the oscillatory errors induced by the time-varying disturbance.

 Overall, these results indicate that explicit modeling and compensaton for periodic disturbances can substantially improve tracking performance compared to the methods assuming static disturbances.

\subsection{Limitations}

The proposed approach was tested in application to the linear disturbances only, leaving rotary for further research.
Since the periodic disturbance estimation relies on the historical data, it takes time to re-evaluate the disturbance after a sudden change in its properties.
The proposed method is not allpicable to stochastic disturbances, where robust MPC or non-periodic adaptive mathods should be used.
The method was shown to perform well in simulation, where the dynamics is modeled in the same way during the whole experiment, and battery discharge, motor heating and sudden surface change cannot happen.
No real-world tests were performed yet, leaving the sim-to-real gap for further investigation.
When no periodic disturbance is applied, the proposed method performs slightly worse than the baseline, making it necessary to evaluate the situation and use one or the other controller in real-world scenarios.

\section{CONCLUSIONS}

An adaptive control strategy for quadrupedal locomotion under periodic external disturbances was proposed.
By extending a baseline MPC framework with a lightweight disturbance regressor, the proposed method can separately estimate and compensate for both stationary and periodic components of unknown forces.
Simulation results suggest enhanced reference tracking performance compared to a standard MPC baseline and a static disturbance compensation approach.

Future work will include comprehensive investigation of torque disturbances, online estimation of varying disturbance frequencies, and real-world hardware experiments. Additional work could also incorporate more complex disturbance models (e.g., stochastic components) and explore methods to accelerate adaptation when environmental changes occur abruptly. Overall, these advances have the potential to make quadrupedal robots more robust and agile in complex, dynamic environments.






\bibliographystyle{IEEEtran}
\bibliography{my-bib}

\end{document}